\documentclass{article} 
\usepackage{tencent_ailab_tech_report}

\usepackage[T1]{fontenc} 
\usepackage[utf8]{inputenc} 

\usepackage{microtype}
\usepackage{hyperref}
\usepackage{url}
\usepackage{booktabs}
\usepackage[table]{xcolor}
\usepackage{colortbl}
\definecolor{darkblue}{rgb}{0, 0, 0.5}
\hypersetup{colorlinks=true, citecolor=darkblue, linkcolor=darkblue, urlcolor=darkblue}

\usepackage{graphicx}
\usepackage{enumitem}
\usepackage{pgfplots}
\usepackage{tabularx}
\usepackage{caption}
\usepackage{subcaption}
\usepackage{bbm}
\usepackage{url}
\usepackage{amsmath} 
\usepackage{pifont}
\usepackage{arydshln}
\usepackage{makecell}
\usepackage{multirow} 
\usepackage{algorithm}
\usepackage{wrapfig}
\usepackage{amssymb}

\usepackage{algpseudocode}

\definecolor{lightgray}{gray}{0.93}
\definecolor{lightblue}{RGB}{220,235,247}

\DeclareUnicodeCharacter{211D}{\ensuremath{\mathbb{R}}}
\DeclareUnicodeCharacter{2124}{\ensuremath{\mathbb{Z}}}
\DeclareUnicodeCharacter{21A6}{\ensuremath{\mapsto}}

\DeclareUnicodeCharacter{2264}{\ensuremath{\leq}}
\DeclareUnicodeCharacter{2265}{\ensuremath{\geq}}
\DeclareUnicodeCharacter{2192}{\ensuremath{\rightarrow}}
\DeclareUnicodeCharacter{2194}{\ensuremath{\leftrightarrow}}
\DeclareUnicodeCharacter{2227}{\ensuremath{\wedge}}
\DeclareUnicodeCharacter{22A2}{\ensuremath{\vdash}}
\DeclareUnicodeCharacter{2080}{\ensuremath{_0}}
\DeclareUnicodeCharacter{2081}{\ensuremath{_1}}
\DeclareUnicodeCharacter{2082}{\ensuremath{_2}}
\DeclareUnicodeCharacter{2083}{\ensuremath{_3}}
\DeclareUnicodeCharacter{2084}{\ensuremath{_4}}
\DeclareUnicodeCharacter{2085}{\ensuremath{_5}}
\DeclareUnicodeCharacter{2086}{\ensuremath{_6}}
\DeclareUnicodeCharacter{2087}{\ensuremath{_7}}
\DeclareUnicodeCharacter{2088}{\ensuremath{_8}}
\DeclareUnicodeCharacter{2089}{\ensuremath{_9}}
\DeclareUnicodeCharacter{2200}{\ensuremath{\forall}}
\DeclareUnicodeCharacter{2203}{\ensuremath{\exists}}
\DeclareUnicodeCharacter{2228}{\ensuremath{\lor}}
\DeclareUnicodeCharacter{2260}{\ensuremath{\neq}}
\DeclareUnicodeCharacter{230B}{\ensuremath{\rfloor}} 
\DeclareUnicodeCharacter{2223}{\ensuremath{\mid}}    
\DeclareUnicodeCharacter{230A}{\ensuremath{\lfloor}} 
\DeclareUnicodeCharacter{2090}{\ensuremath{_a}}  
\DeclareUnicodeCharacter{2091}{\ensuremath{_e}}  
\DeclareUnicodeCharacter{2092}{\ensuremath{_o}}  
\DeclareUnicodeCharacter{2093}{\ensuremath{_x}}  
\DeclareUnicodeCharacter{2211}{\ensuremath{\sum}}        
\DeclareUnicodeCharacter{2115}{\ensuremath{\mathbb{N}}}  

\DeclareUnicodeCharacter{2208}{\ensuremath{\in}} 
\title{\textsc{Clue}: Non-parametric Verification from Experience via Hidden-State Clustering}

\author{
Zhenwen Liang$^{1,}$$^\dagger$,
Ruosen Li$^{1,2,^\dagger,}$\thanks{Work done during Ruosen's and Yujun's Internship at Tencent AI Lab.}\;, 
Yujun Zhou$^{1,3,*}$,
\textbf{
Linfeng Song$^1$,
Dian Yu$^1$,
Xinya Du$^2$,}\\
\textbf{
Haitao Mi$^1$,
Dong Yu$^1$}
\vspace{1em}\\
$^1$Tencent AI Lab, 
$^2$University of Texas at Dallas,
$^3$University of Notre Dame
\\
$^\dagger$ Equal contribution \quad
\\
Correspondence to: \texttt{zhenwzliang@global.tencent.com}
}

\colmfinalcopy 
\begin{document}

\maketitle

\begin{abstract}
Assessing the quality of Large Language Model (LLM) outputs presents a critical challenge. Previous methods either rely on text-level information (e.g., reward models, majority voting), which can overfit to superficial cues, or on calibrated confidence from token probabilities, which would fail on less-calibrated models. Yet both of these signals are, in fact, partial projections of a richer source of information: the model’s internal hidden states. Early layers, closer to token embeddings, preserve semantic and lexical features that underpin text-based judgments, while later layers increasingly align with output logits, embedding confidence-related information. This paper explores hidden states directly as a unified foundation for verification. We show that the correctness of a solution is encoded as a geometrically separable signature within the trajectory of hidden activations. To validate this, we present \textbf{\textsc{Clue} (Clustering and Experience-based Verification)}, a deliberately minimalist, non-parametric verifier. With no trainable parameters, \textsc{Clue} only summarizes each reasoning trace by a hidden state delta and classifies correctness via nearest-centroid distance to "success" and "failure" clusters formed from past experience. The simplicity of this method highlights the strength of the underlying signal. Empirically, \textsc{Clue} consistently outperforms LLM-as-a-judge baselines and matches or exceeds modern confidence-based methods in reranking candidates, improving both top-1 and majority-vote accuracy across AIME 24/25 and GPQA. As a highlight, on AIME 24 with a 1.5B model, \textsc{Clue} boosts accuracy from 56.7\% (majority@64) to 70.0\% (top-maj@16). 
\end{abstract}


\section{Introduction}
The remarkable ability of Large Language Models (LLMs) to generate numerous potential solutions for complex problems has also created a difficult new challenge: verification \citep{cobbe2021training,lightman2023let,hosseini2024v}. When a model produces dozens of different, plausible-looking answers for a single prompt, the task is no longer just about generation. It becomes a critical problem of selection: how can we reliably find the single correct answer within a flood of convincing but incorrect alternatives? 

\begin{figure}
\centering
\includegraphics[width=\linewidth]{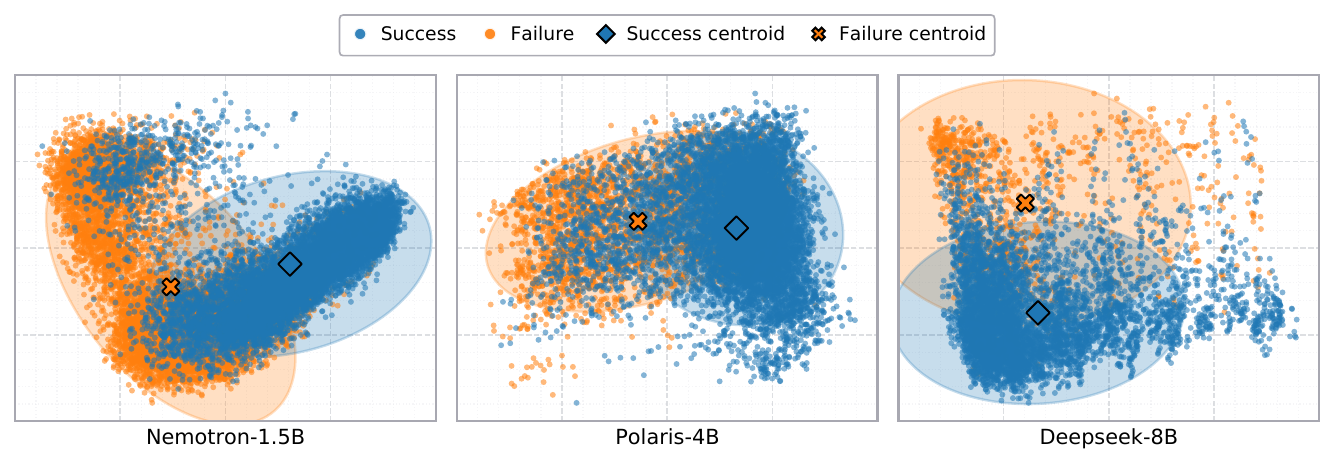}
\caption{Visualization of hidden state trajectories for correct (blue) and incorrect (orange) solutions from our experience set, projected to 2D using PCA. Each panel displays data from a different base model. Across all models, a geometric separation is visible.}
\vspace{-0.4cm}
\label{fig:intro_clusters}
\end{figure}

To address this question, the research community has largely pursued two main strategies. The first operates purely on the surface of the generated text, delegating evaluation to an external judge. This includes training separate reward models \citep{ouyang2022training,bai2022constitutional,zheng2023judging} or adopting simple heuristics such as majority voting \citep{wang2022self}. While useful in practice, these after-the-fact approaches are fundamentally blind to the model’s actual reasoning process. They can be systematically misled by stylistic artifacts—e.g., verbose but incorrect answers often receive higher scores than terse but correct ones \citep{glaese2022improving}. Moreover, trained judges inherit biases and limitations from their training data, making them brittle under distribution shift and expensive to retrain for new domains.

A second line of work attempts to go beneath the surface, but only through the shadow of the model’s reported confidence. Methods in this category rely on token probabilities, entropy, or derived uncertainty estimates \citep{kadavath2022language,lin2023generating,geng2023survey,xiong2024efficient,fu2025deep}. The underlying assumption is that higher probability correlates with higher correctness. However, calibration of LLMs remains poor: even state-of-the-art models are often “confidently wrong,” assigning extreme likelihood to factually false or logically inconsistent outputs \citep{fu2025multiple}. These confidence-based metrics also degrade significantly on smaller or less-tuned models, where probability distributions are noisier and less interpretable, making them a fragile basis for reliable verification.

In this work, we argue that correctness is most faithfully captured not in the final text nor solely in surface-level confidence, but in the model’s internal reasoning trajectory itself. Hidden states naturally subsume both kinds of information: earlier layers, being closer to the token embeddings, encode rich semantic and lexical features that underlie text-based judgments, while later layers progressively align with the output logits, embedding signals that correlate with confidence and probability. In this sense, hidden states offer a unified and more fundamental representation that integrates both semantic and probabilistic cues. Our core hypothesis is that within this trajectory, correctness manifests as a geometrically separable pattern that distinguishes success from failure. Crucially, this separation is not just theoretical but empirically observable: as illustrated in Figure~\ref{fig:intro_clusters}, trajectories for correct and incorrect solutions consistently cluster apart across tasks and model scales, echoing insights from mechanistic interpretability \citep{nostalgebraist2020interpreting,belrose2023eliciting,tomaz2025momentum} while extending them toward actionable verification.

This visually evident structure in hidden-state space motivates a deliberately lightweight verifier. If correct and incorrect trajectories occupy separable regions, then a complex learned judge may be unnecessary. We introduce \textbf{\textsc{Clue} (Clustering and Experience-based Verification)}, a training-free, supervised aggregation framework that operates directly on the model’s internal computation. Rather than using absolute final states, \textsc{Clue} summarizes each reasoning trace by an activation delta—the difference between hidden states at the start and end of the explicit reasoning block (delimited by <think> … </think>). Intuitively, this delta factors out prompt conditioning and isolates the transformation induced by reasoning itself. From labeled historical trajectories, \textsc{Clue} computes two reference centroids—one for successes and one for failures—and classifies a new trace by its proximity to these centroids (using a layer-averaged Euclidean distance). 

Our experiments support this premise. \textsc{Clue} consistently matches or surpasses strong LLM-as-a-judge and confidence-based baselines, with especially clear gains on smaller or less-calibrated models where probability cues are unreliable. Because \textsc{Clue} performs a one-time, deterministic aggregation without gradient training, it avoids many overfitting failure modes of learned verifiers and exhibits robust generalization across tasks and model scales. More broadly, these results provide evidence that the hidden states of LLMs encode a rich, structured signal for verification.

\section{Related Work}

\subsection{Latent Reasoning and Activation Geometry}
LLMs can reason in latent space instead of (or alongside) explicit token chains, via continuous “thought states” fed back into the model or compact hidden “thinking tokens” that compress CoT~\citep{hao2024training,shen2025efficient}. Interpretability tools like the logit lens and tuned lens show that intermediate activations progressively align with output distributions, suggesting layer-wise decodable semantics and confidence signals~\citep{nostalgebraist2020interpreting,belrose2023eliciting}. Hidden-state probes can \emph{self-verify} intermediate answers and enable early exit~\citep{zhang2025reasoningmodelsknowtheyre}, while semantic clustering of hidden rationales can improve robustness~\citep{knappe2024semantic}. Beyond verification, activation directions can monitor or steer model traits (e.g., sycophancy, hallucination) via \emph{persona vectors}~\citep{chen2025personavectorsmonitoringcontrolling}. Also, in-context activation vectors indicate that linear structure in hidden space can be mapped and reused across tasks~\citep{liu2024incontextvectorsmakingcontext}. More broadly, recent surveys on representation engineering highlight how linear directions and activation editing provide a general lens on hidden-state geometry in LLMs~\citep{bartoszcze2025representationengineeringlargelanguagemodels}.
 Unlike these trained probes or steering methods, our verifier \textsc{Clue} is training-free and purely reads cross-layer activation deltas.

\subsection{Test Time Scaling}
Recent research has increasingly focused on test-time scaling – techniques that improve model performance by allocating more computation during inference without changing the model’s parameters.
Parallel approaches (such as self-consistency~\citep{wang2022self} and ensemble “best-of-N” selection~\citep{snell2024scaling}) generate multiple independent chain-of-thought solutions and then aggregate or vote on the final answer, significantly boosting accuracy on complex tasks.
Sequential approaches (such as iterative self-refinement~\citep{madaan2023self}, Tree-of-Thoughts search~\citep{yao2023tree}) allow the model to think in multiple steps, using intermediate reasoning to inform subsequent generations. Variants like weighted or semantic self-consistency~\citep{luo2024improvemathematicalreasoninglanguage,knappe2024semantic} highlight the importance of aggregating diverse rationales, while RLHF and LLM-as-a-judge approaches~\citep{ouyang2022training,zheng2023judging} provide external supervision but can be costly and biased. To reduce dependence on large reward models, SWIFT learns \emph{lightweight} hidden-state rewards that scale efficiently to best-of-\textit{N} sampling~\citep{guo2025miningintrinsicrewardsllm}. Complementary to this, DeepConf filters low-quality reasoning traces using internal confidence signals, improving both efficiency and accuracy~\citep{fu2025deep}; relatedly, early-exit schemes can truncate overthinking while preserving accuracy~\citep{kadavath2022languagemodelsmostlyknow,yang2025dynamicearlyexitreasoning}. In contrast, \textsc{Clue} introduces no trainable verifier: it computes success/failure centroids once from past experience and reranks by nearest centroid, showing that correctness is geometrically separable in hidden space.

\section{The \textsc{Clue} Framework}
\label{sec:methodology}

\begin{figure}[ht]
    \centering
    \includegraphics[width=0.88\linewidth]{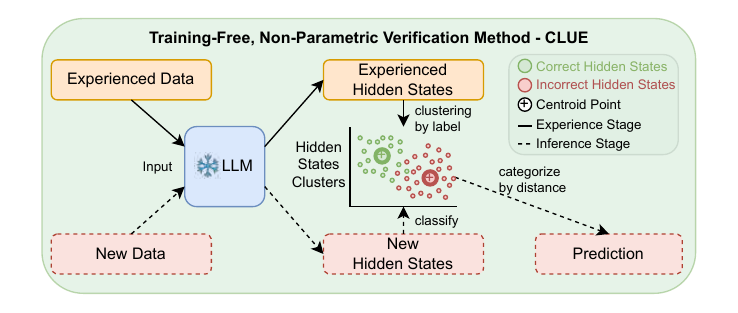}
    \caption{Overview of \textsc{Clue}. \textbf{Left (Learning):} Labeled historical trajectories are summarized by their activation deltas and aggregated into success and failure centroid matrices. \textbf{Right (Verification):} A new trajectory is summarized by its activation delta and classified by the layer-averaged Euclidean distance (Eq.~\ref{eq:layeravg}) to the two pre-computed centroids. The underlying LLM remains \emph{frozen} throughout.}
    \label{fig:methodology}
\end{figure}

We first outline the core intuition behind \textsc{Clue}. Each time an LLM solves a problem, its internal computation traces a trajectory through a high-dimensional representation space. We hypothesize that trajectories leading to correct solutions differ systematically from those leading to incorrect ones. \textsc{Clue} captures this difference via a training-free, supervised aggregation over activation deltas: it summarizes each labeled trajectory, builds class centroids from these summaries, and then verifies a new trajectory by proximity to the centroids. This section formalizes the setup and the resulting geometric rule.

\subsection{Problem Formulation}
Let an LLM be tasked with generating a response $R_i$ for a prompt $P$. We define a \emph{trajectory} (or \emph{experience}) $T_i=(P,R_i)$, paired with a ground-truth binary label $y_i\in\{0,1\}$, where $y_i=1$ denotes a correct solution (success) and $y_i=0$ denotes an incorrect solution (failure). The goal is to learn a verification function $f$ that maps a new trajectory $T_{\text{new}}$ to a prediction $\hat{y}_{\text{new}}=f(T_{\text{new}})\in\{0,1\}$. Unlike text-based or probability-based approaches, $f$ operates exclusively on the hidden-state representations generated during the production of $R_{\text{new}}$. The LLM parameters are kept fixed (frozen) throughout both learning and inference.

\subsection{\textsc{Clue}: Verification via Activation-Delta Summaries}
\label{sec:delta}
The central hypothesis is that the \emph{transformation} of internal states during explicit reasoning contains a robust signal of correctness. We capture this transformation with an \emph{activation delta}, defined as the difference between hidden states at the start and the end of the reasoning block. In our experiments, the reasoning block is delimited by \texttt{<think>} and \texttt{</think>}.

Let the model have $L$ layers and hidden dimension $D$. For a given trajectory $T$, denote by
\[
\mathbf{h}_{\text{start}}(T)\in\mathbb{R}^{L\times D}
\quad\text{and}\quad
\mathbf{h}_{\text{end}}(T)\in\mathbb{R}^{L\times D}
\]
the matrices of hidden states extracted, respectively, at the final token of \texttt{<think>} (just before detailed reasoning) and at the final token of \texttt{</think>} (after the reasoning has been formed). The activation delta is the matrix
\begin{equation}
\Delta \mathbf{h}(T) \;=\; \mathbf{h}_{\text{end}}(T)\;-\;\mathbf{h}_{\text{start}}(T)
\;\in\;\mathbb{R}^{L\times D}.
\label{eq:delta}
\end{equation}
We use hidden states from \emph{all} layers, reflecting the assumption that correctness-related information is distributed across depth (earlier layers retain semantic/lexical cues; later layers align more strongly with logits). The activation-delta matrix $\Delta \mathbf{h}(T)$ serves as the sole feature representation for verification.

\subsection{Centroid Construction and Classification}
\label{sec:centroids}
The learning phase is a one-time deterministic statistical aggregation over labeled trajectories $\mathcal{D}=\{(T_i,y_i)\}_{i=1}^N$. Define index sets
\[
\mathcal{I}_{\text{succ}}=\{\,i\mid y_i=1\,\},\qquad
\mathcal{I}_{\text{fail}}=\{\,i\mid y_i=0\,\}.
\]
For each trajectory, compute its activation delta $\Delta \mathbf{h}_i=\Delta \mathbf{h}(T_i)$ as in Eq.~\eqref{eq:delta}. The success and failure \emph{centroid matrices} are the element-wise means:
\begin{equation}
\mathbf{V}_{\text{succ}} \;=\; \frac{1}{|\mathcal{I}_{\text{succ}}|}\sum_{i\in \mathcal{I}_{\text{succ}}}\Delta \mathbf{h}_i,
\qquad
\mathbf{V}_{\text{fail}} \;=\; \frac{1}{|\mathcal{I}_{\text{fail}}|}\sum_{i\in \mathcal{I}_{\text{fail}}}\Delta \mathbf{h}_i.
\label{eq:centroids}
\end{equation}
Both $\mathbf{V}_{\text{succ}},\mathbf{V}_{\text{fail}}\in\mathbb{R}^{L\times D}$ are stored for inference.

At inference, a new trajectory $T_{\text{new}}$ is summarized by $\Delta \mathbf{h}_{\text{new}}=\Delta \mathbf{h}(T_{\text{new}})$. We compare it to the two centroids using the \emph{layer-averaged Euclidean distance}. For two matrices $\mathbf{A},\mathbf{B}\in\mathbb{R}^{L\times D}$ with row vectors $\mathbf{a}_l,\mathbf{b}_l\in\mathbb{R}^{D}$ (the $l$-th layer representations), define
\begin{equation}
d(\mathbf{A},\mathbf{B})
\;=\;
\frac{1}{L}\sum_{l=1}^{L}\bigl\|\mathbf{a}_l-\mathbf{b}_l\bigr\|_2.
\label{eq:layeravg}
\end{equation}
Compute $d_{\text{succ}}=d(\Delta \mathbf{h}_{\text{new}},\mathbf{V}_{\text{succ}})$ and
$d_{\text{fail}}=d(\Delta \mathbf{h}_{\text{new}},\mathbf{V}_{\text{fail}})$, and classify as
\[
\hat{y}_{\text{new}} \;=\;
\begin{cases}
1, & \text{if } d_{\text{succ}}<d_{\text{fail}},\\[2pt]
0, & \text{otherwise.}
\end{cases}
\]
This rule matches the high-level description in Figure~\ref{fig:methodology} and requires no gradient-based optimization.

\subsection{Application to Solution Reranking}
\label{sec:rerank}
The geometric formulation provides a continuous quality score that is naturally suited for reranking. Given a prompt $P$ and $k$ responses $\{R_1,\dots,R_k\}$, form trajectories $\{T_1,\dots,T_k\}$ and their activation deltas $\{\Delta \mathbf{h}_1,\dots,\Delta \mathbf{h}_k\}$. Define for each candidate
\begin{equation}
s_j \;=\; d\!\bigl(\Delta \mathbf{h}_j,\mathbf{V}_{\text{succ}}\bigr),
\qquad \text{(lower is better)}
\label{eq:score}
\end{equation}
and rank candidates in ascending order of $s_j$. This ranking can be used for top-1 selection or to improve aggregation schemes such as majority vote by prioritizing candidates whose internal reasoning is closest to the success centroid.

\subsection{Rationale for a Minimalist, Experience-Based Design}
\label{sec:rationale}
\textsc{Clue} is intentionally minimalist to isolate the contribution of the representation itself. If a simple, training-free geometric rule over activation-delta summaries yields strong verification performance, this provides evidence that correctness signals are geometrically encoded and separable in activation space. In addition, the one-time deterministic aggregation in Eq.~\eqref{eq:centroids} reduces the risk of overfitting associated with learned verifiers and supports robust generalization across tasks and model scales. By leveraging the geometry of \emph{how} solutions are computed, \textsc{Clue} offers a lightweight and broadly applicable path for verification that complements text-level and confidence-based signals.

\section{Experiments}
\label{sec:experiments}

To rigorously evaluate the effectiveness of our non-parametric, hidden-state-based verifier, we designed a series of experiments targeting both in-domain mathematical reasoning and out-of-distribution general reasoning tasks. Our evaluation is structured around two primary objectives: first, to assess the raw classification accuracy of our method in distinguishing correct from incorrect solutions, and second, to measure its ability to improve overall reasoning performance by reranking multiple candidate solutions.

\subsection{Datasets and Model Configuration}
Our methodology relies on an "experience set" to establish the geometric reference points for successful and failed reasoning. For this purpose, we curated a comprehensive collection of mathematical problems from two standard benchmarks: AIME (from 1983 to 2023) \citep{aime_1983_2024} and the MATH \citep{hendrycks2021measuring} dataset (specifically, problems of level 3 to 5). These datasets provide a diverse and challenging foundation for learning the characteristic activation patterns of mathematical reasoning. To test the performance and generalization of our approach, we use three distinct test sets. For in-domain evaluation, we use AIME 2024 and AIME 2025, which follow the same distribution as our experience data. To assess out-of-distribution (OOD) generalization, we evaluate on GPQA \citep{rein2024gpqa}, a benchmark focused on graduate-level questions that demand complex, general reasoning abilities beyond the mathematical domain.

Our experiments cover a range of model sizes and architectures to ensure our findings are not specific to a single model's capabilities. We selected three distinct reasoning models: \textbf{Nemotron-Research-Reasoning-Qwen-1.5B} \citep{liu2025prorl}, a smaller yet capable model; \textbf{Polaris-4B} \citep{Polaris2025}, a mid-sized model; and \textbf{DeepSeek-R1-0528-Qwen3-8B} \citep{guo2025deepseek}, a larger and more powerful model. To test the sensitivity of our method to the length and complexity of the reasoning trace, we conducted experiments with varied generation lengths for each model, specifically 16k, 32k, and 64k tokens. We use recommended inference setting including temperature, system prompt, from their Huggingface repository.

The process for constructing the experience set was as follows: for each problem in the AIME and MATH datasets, we sampled 32 unique solutions from the respective model. Each generated solution was then evaluated using a deterministic, rule-based verifier to obtain a ground-truth label of correct or incorrect. Then, we randomly selected 10,000 correct and 10,000 incorrect trajectories to form a balanced experience set. This set was used to compute the success and failure centroids. For the evaluation phase, we generated 64 candidate solutions for each problem in our test sets.

\subsection{Evaluation Setups and Baselines}
We evaluate our method, which we refer to as \textbf{\textsc{Clue}}, across two distinct experimental setups.

The first setup frames the task as a binary classification problem to directly measure the verifier's accuracy. For each of the 64 sampled solutions on the test sets, our \textsc{Clue} method predicts a label of correct or incorrect based on whether the solution's activation delta is closer to the success or failure centroid. The ground truth for this task is again determined by the rule-based verifier. We compare our method against several strong baselines. We compare against powerful LLMs serving as judges. Specifically, we use \textbf{GPT-4o} \citep{hurst2024gpt} in an LLM-as-a-judge capacity. We evaluate the judge in two settings to control for the information they can access: one where the full solution, including the entire \texttt{<think>} block, is provided to the LLM judge, and another where only the part after the thinking process is provided. The former tests the judge's ability to evaluate the reasoning process, while the latter tests its ability to verify the result itself.

\begin{table}
\centering
\caption{Binary classification performance of different verifiers on solutions generated by Nemotron-1.5B and Polaris-4B. Our method (\textsc{Clue}) is compared against an LLM-as-a-judge baseline. We report overall Accuracy, True Positive Rate (TPR), and True Negative Rate (TNR).}
\label{tab:classification_results}
\renewcommand{\arraystretch}{1.25} 
\resizebox{\textwidth}{!}{
\begin{tabular}{l ccc ccc}
\toprule
& \multicolumn{3}{c}{\textbf{Nemotron-1.5B Solutions}} & \multicolumn{3}{c}{\textbf{Polaris-4B Solutions}} \\
\cmidrule(lr){2-4} \cmidrule(lr){5-7} 
\textbf{Verifier Method} & \textbf{Accuracy (\%)} & \textbf{TPR (\%)} & \textbf{TNR (\%)} & \textbf{Accuracy (\%)} & \textbf{TPR (\%)} & \textbf{TNR (\%)} \\
\midrule
\multicolumn{7}{l}{\textit{Test Set: AIME 2024}} \\
\quad \textbf{\textsc{Clue} (Ours)} & \textbf{80.9} & 72.9 & 87.4 & \textbf{81.1} & 89.5 & 51.3 \\
\quad GPT-4o (Answer Only) & 58.6 & 57.2 & 59.7 & 80.1 & 84.3 & 63.1 \\
\quad GPT-4o (Full Trace) & 47.5 & 45.8 & 48.2 & 64.3 & 70.9 & 50.2 \\
\midrule
\multicolumn{7}{l}{\textit{Test Set: AIME 2025}} \\
\quad \textbf{\textsc{Clue} (Ours)} & \textbf{85.2} & 82.9 & 86.4 & \textbf{77.7} & 80.7 & 70.1 \\
\quad GPT-4o (Answer Only) & 59.2 & 60.8 & 58.3 & 73.0 & 85.1 & 34.4 \\
\quad GPT-4o (Full Trace) & 47.1 & 48.6 & 46.7 & 59.3 & 69.8 & 27.6 \\
\bottomrule
\end{tabular}}
\end{table}

The second setup evaluates the practical impact of our method on improving final reasoning accuracy through reranking. Here, for each test problem, we use \textsc{Clue} to rerank the 64 generated solutions. The ranking criterion is the Euclidean distance of a solution's activation delta to the success centroid, with smaller distances indicating higher quality. We report our performance using several metrics: \textbf{top@1}, the accuracy of the single best-ranked solution; and \textbf{top-maj@k}, the accuracy achieved by performing majority voting on the answers from the top-\textit{k} ranked solutions, for $k \in \{4, 8, 16\}$. We compare these results against a suite of standard and state-of-the-art baselines. These include \textbf{mean@64}, which measures the average accuracy of a single randomly sampled solution; \textbf{majority@64}, the accuracy of standard majority voting over all 64 samples; \textbf{DeepConf@64} \citep{fu2025deep}, a recent and competitive method that uses model confidence scores for reranking; and \textbf{pass@64}, which represents the oracle upper bound, indicating whether at least one correct answer exists among the 64 samples.

\subsection{Classification Performance}

We first evaluate our method, \textsc{Clue}, on its core capability: accurately classifying individual solutions as either correct or incorrect. Table~\ref{tab:classification_results} presents the performance of our verifier compared to a strong LLM-as-a-judge baseline (GPT-4o) on solutions generated by both a smaller model (Nemotron-1.5B) and a more capable one (Polaris-4B). We report overall accuracy, as well as the True Positive Rate (TPR), which measures the ability to correctly identify successful solutions, and the True Negative Rate (TNR), which measures the ability to correctly identify failed solutions.

The results clearly indicate that our \textsc{Clue} verifier provides a substantial and consistent advantage over the LLM-as-a-judge baseline. A key observation is that the LLM judge exhibits a strong optimistic bias, frequently misclassifying incorrect solutions as correct. This is evident in its consistently low True Negative Rate, which drops to a mere 34.4\% for Polaris-4B solutions on AIME 2025. This inherent optimism explains why the LLM judge's performance appears to improve significantly when evaluating solutions from a stronger base reasoner like Polaris-4B. A more capable reasoner produces a higher proportion of correct solutions, which the LLM judge identifies with reasonable accuracy (high TPR). Consequently, the judge's primary weakness—its failure to reliably identify incorrect answers—has a diminished impact on its overall accuracy score simply because there are fewer negative samples to misclassify. In contrast, our \textsc{Clue} method demonstrates a more robust and balanced performance profile. It maintains a very high TNR (up to 87.4\%) when evaluating the weaker Nemotron-1.5B model, making it highly effective at filtering out the larger volume of incorrect attempts. Simultaneously, it achieves a high TPR on outputs from the stronger Polaris-4B model (up to 89.5\%), showing it is equally adept at recognizing valid reasoning. This balanced capability makes our approach a more universally effective verifier, providing significant benefits regardless of the base model's reasoning proficiency.

\begin{table}
\centering
\caption{Reasoning accuracy on AIME and GPQA test sets after reranking 64 candidate solutions. Results are presented as percentages (\%). `mean@64` represents the average accuracy of a single sample, while `pass@64` is the oracle upper bound.}
\label{tab:reranking_results}
\renewcommand{\arraystretch}{1.25} 
\resizebox{\textwidth}{!}{
\begin{tabular}{l ccc ccc ccc}
\toprule
& \multicolumn{3}{c}{\textbf{Nemotron-1.5B}} & \multicolumn{3}{c}{\textbf{Polaris-4B}} & \multicolumn{3}{c}{\textbf{DeepSeek-8B}} \\
\cmidrule(lr){2-4} \cmidrule(lr){5-7} \cmidrule(lr){8-10} 
\textbf{Metric} & \textbf{AIME 24} & \textbf{AIME 25} & \textbf{GPQA} & \textbf{AIME 24} & \textbf{AIME 25} & \textbf{GPQA} & \textbf{AIME 24} & \textbf{AIME 25} & \textbf{GPQA} \\
\midrule
\multicolumn{10}{l}{\textit{Baselines}} \\
\quad mean@64 & 45.0 & 35.0 & 41.9 & 79.2 & 75.4 & 55.2 & 87.1 & 75.8 & 54.86 \\
\quad majority@64 & 56.7 & 36.7 & 44.4 & 80.0 & 80.0 & 56.6 & 90.0 & 83.3 & 61.11 \\
\quad DeepConf@64 & 56.7 & 30.0 & 40.2 & 80.0 & 73.3 & 55.7 & \textbf{93.3} & \textbf{86.7} & 62.12 \\
\quad pass@64 (Oracle) & 76.7 & 63.3 & 83.8 & 86.7 & 90.0 & 88.4 & 93.3 & 93.3 & 94.85 \\
\midrule
\multicolumn{10}{l}{\textit{\textsc{Clue} Reranking (Ours)}} \\
\quad top@1 & 66.7 & 40.0 & 46.5 & \textbf{83.3} & 76.7 & 52.5 & 90.0 & 83.3 & 56.57 \\
\quad top-maj@4 & \textbf{70.0} & 40.0 & 43.9 & \textbf{83.3} & 76.7 & 57.1 & 90.0 & \textbf{86.7} & 61.62 \\
\quad top-maj@8 & \textbf{70.0} & 40.0 & \textbf{47.0} & 80.0 & 80.0 & 58.1 & \textbf{93.3} & \textbf{86.7} & 61.11 \\
\quad top-maj@16 & \textbf{70.0} & \textbf{43.3} & 44.4 & 80.0 & \textbf{83.3} & \textbf{59.6} & \textbf{93.3} & \textbf{86.7} & \textbf{62.63} \\
\bottomrule
\end{tabular}}
\end{table}

\subsection{Reranking for Enhanced Reasoning Accuracy}

Moving beyond binary classification, we evaluate \textsc{Clue} as a reranking tool to improve reasoning accuracy. By scoring and reordering 64 candidate solutions per problem, \textsc{Clue} consistently outperforms majority voting on both in-domain AIME and out-of-domain GPQA benchmarks (Table~\ref{tab:reranking_results}). For example, with Nemotron-1.5B on AIME 24, ``top-maj@16'' reaches 70.0\% versus 56.7\% for ``majority@64.'' Even ``top@1'' often surpasses majority voting, showing the effectiveness of \textsc{Clue}’s distance-based scoring.  

This advantage extends to general reasoning: on GPQA, Polaris-4B achieves 59.6\% with \textsc{Clue} versus 56.6\% with majority voting. Such transfer demonstrates that the geometric separation of success and failure in hidden states reflects a fundamental property of reasoning, not domain-specific patterns. Compared with the confidence-based baseline DeepConf, \textsc{Clue} exhibits greater robustness. While both methods perform strongly on DeepSeek-8B, DeepConf collapses on weaker models, often below majority voting. \textsc{Clue}, however, maintains its edge across all scales, leveraging internal reasoning signals that remain geometrically separable even when output confidences are poorly calibrated. This highlights \textsc{Clue}’s broad applicability, particularly for smaller models where confidence cues are unreliable.

\subsection{Generalization and the Influence of Training Paradigms}

\begin{figure*}
    \centering
    \includegraphics[width=\textwidth]{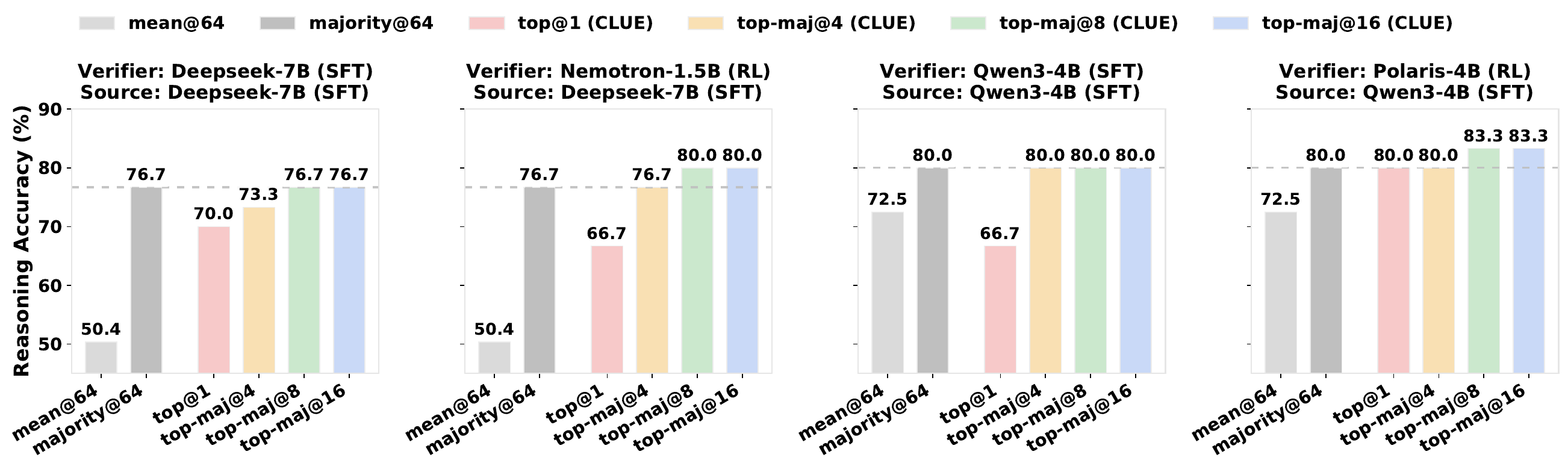}
    \caption{Cross-model reranking performance on AIME 24. The results show that RL-trained models (Nemotron-1.5B, Polaris-4B) are not only effective at self-verification but are also superior verifiers for trajectories generated by SFT-trained models (Deepseek-7B, Qwen3-4B).}
    \label{fig:cross_model_results}
\end{figure*}

We next examine \textsc{Clue}’s behavior across training paradigms and models. Our hypothesis is that the geometric separability of success and failure in hidden states depends strongly on training methodology---specifically, the contrast between Supervised Fine-Tuning (SFT) and Reinforcement Learning (RL). We evaluated four models: two SFT/distillation-based (\textbf{Deepseek-7B} \citep{guo2025deepseek}, \textbf{Qwen3-4B} \citep{yang2025qwen3}) and two RL-tuned (\textbf{Nemotron-1.5B}, \textbf{Polaris-4B}). In a cross-model setup, reasoning traces from one model were fed into another’s hidden states for reranking, enabling both self- and cross-verification tests.  

As shown in Figure~\ref{fig:cross_model_results}, SFT models struggle: their self-reranking (``top-maj@16'') barely matches or even lags the ``majority@64'' baseline, indicating weak internal separation of correctness. By contrast, RL models act as strong verifiers even across models: Nemotron-1.5B boosts Deepseek-7B’s accuracy to 80.0\% (vs 76.7\% baseline), and Polaris-4B lifts Qwen3-4B’s outputs to 83.3\% (vs 80.0\% self-rerank).

We attribute this gap to training signals. SFT trains imitation of correct paths but lacks explicit negative feedback, leaving ``wrongness'' underrepresented. RL, especially with verifiable rewards, supplies direct contrastive supervision, producing geometrically distinct clusters for correct vs.\ incorrect reasoning. This makes RL-trained models inherently stronger verifiers, both for themselves and others.  
\begin{table}
\centering
\caption{Binary classification performance on the general-purpose WebInstruct-verified dataset. We compare \textsc{Clue}'s accuracy against a GPT-4o judge on solutions generated by 1.5B and 4B models. The centroids for \textsc{Clue} were computed using the WebInstruct \citep{ma2025general} training set.}
\label{tab:webinstruct_results}
\renewcommand{\arraystretch}{1.25} 
\begin{tabular}{l cc}
\toprule
& \multicolumn{2}{c}{\textbf{{Reasoner Model}}} \\
\cmidrule(lr){2-3} 
\textbf{Verifier Method} & \textbf{Nemotron-1.5B} & \textbf{Polaris-4B} \\
\midrule
Test Set Composition (Success / Failure) & 1,263 / 2,737 & 1,584 / 2,024 \\
\midrule
\textbf{\textsc{Clue} (Ours)} & \textbf{60.4\%} & \textbf{59.2\%} \\
GPT-4o (LLM-as-a-judge) & 54.0\% & 48.1\% \\
\bottomrule
\end{tabular}
\end{table}

\subsection{Generalization to Diverse, Non-Mathematical Reasoning}
To test \textsc{Clue}’s generalization beyond mathematics, we evaluated it on the diverse \textbf{WebInstruct-verified} benchmark, which spans physics, law, finance, and the humanities. Centroids were built from 5k training questions with generated solutions, and evaluation was conducted on 1k test questions. Ground-truth correctness labels were obtained by providing reference answers to GPT-4o, while GPT-4o itself—without access to the reference—served as the LLM-as-a-judge baseline.

As shown in Table~\ref{tab:webinstruct_results}, \textsc{Clue} consistently outperforms GPT-4o across both 1.5B and 4B models. On the 1.5B model, \textsc{Clue} reaches 60.4\% accuracy versus GPT-4o’s 54.0\%. Most notably, for the 4B model, the LLM judge collapses to 48.1\% (below random), while \textsc{Clue} maintains 59.2\%.

These results provide strong evidence that correctness signals are encoded geometrically in hidden-state trajectories even outside mathematics. Unlike surface-level textual judgments, which fail in heterogeneous domains, \textsc{Clue} extracts a more stable and transferable representation of success versus failure, underscoring its robustness as a general-purpose verifier.

\subsection{Layer-wise Separability Analysis}
Next, we analyze the layer-wise structure of activation-delta matrices to visualize and quantify how class separability emerges from shallow to deep layers.

\begin{figure}
    \centering
    \includegraphics[width=\linewidth]{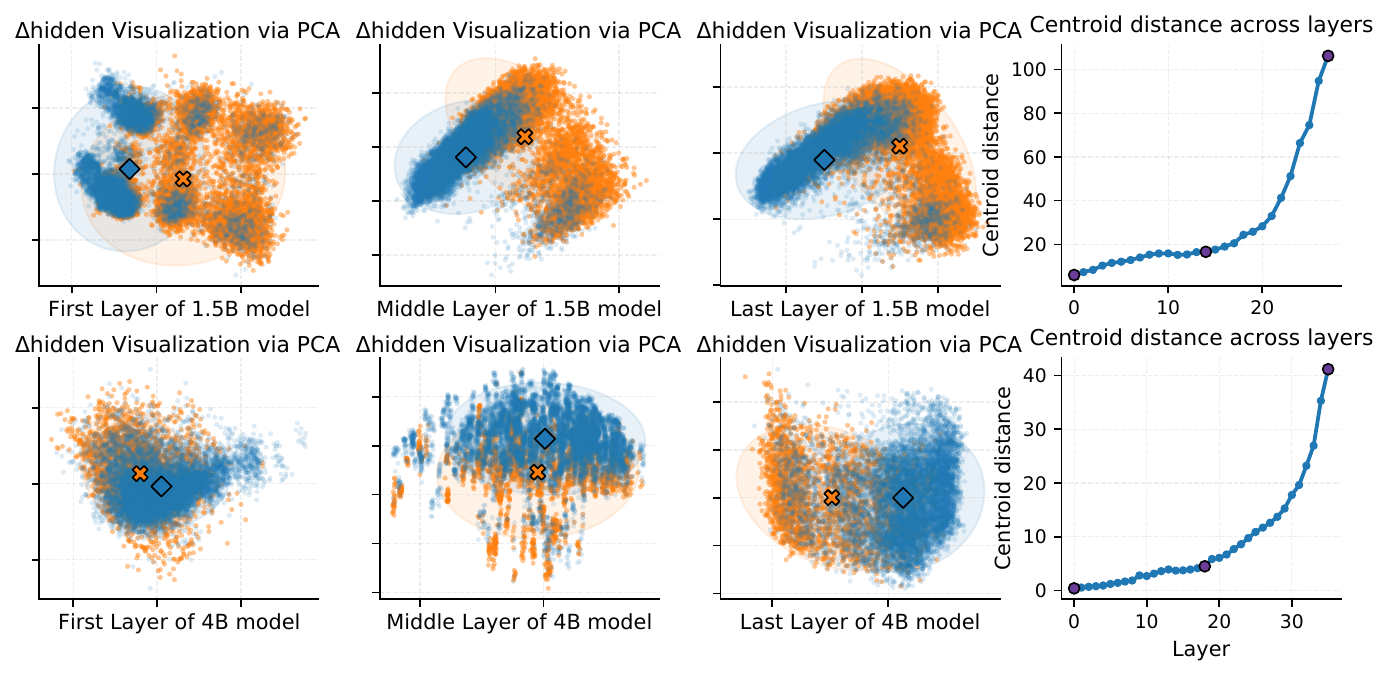}
    \vspace{-0.8em}
    \caption{Layer-wise separability. Each row shows PCA projections from a shallow, a middle, and the final layer, plus a curve of the centroid distance $d^{(\ell)}$ across all layers. The centroid-distance curve increases with $\ell$, indicating stronger correctness signals at deeper layers.}
    \vspace{-1.1em}
    \label{fig:layerwise-visualization}
    
\end{figure}

\textbf{Visualization.}
We project layer-specific activation deltas onto two principal components via PCA. For a trajectory $i$ and layer $\ell$, let $\Delta \mathbf{h}_i^{(\ell)}\in\mathbb{R}^{D}$ denote the $\ell$-th row of $\Delta \mathbf{h}_i\in\mathbb{R}^{L\times D}$. We select representative shallow, middle, and final layers, apply PCA to $\{\Delta \mathbf{h}_i^{(\ell)}\}$, and plot the resulting 2D projections for successes and failures. As shown in the first three columns of each row in Figure~\ref{fig:layerwise-visualization}, the classes are largely overlapping in shallow layers, begin to separate in middle layers, and form compact, well-defined clusters in the final layers.

\textbf{Quantification.}
Let $\mathcal{I}_{\text{succ}}$ and $\mathcal{I}_{\text{fail}}$ be the index sets defined in \S\ref{sec:centroids}. For each layer $\ell\in\{1,\dots,L\}$, we compute layer-wise centroid by averaging the corresponding rows of the activation-delta matrices:
\begin{align}
\mathbf{V}_{\text{succ}}^{(\ell)} = \frac{1}{|\mathcal{I}_{\text{succ}}|} \sum_{i\in\mathcal{I}_{\text{succ}}} \Delta \mathbf{h}_i^{(\ell)} \in \mathbb{R}^{D}, 
\qquad
\mathbf{V}_{\text{fail}}^{(\ell)} = \frac{1}{|\mathcal{I}_{\text{fail}}|} \sum_{i\in\mathcal{I}_{\text{fail}}} \Delta \mathbf{h}_i^{(\ell)} \in \mathbb{R}^{D}.
\end{align}
We then measure the Euclidean distance between the two centroids at layer $\ell$:
\begin{equation}
d^{(\ell)} \;=\; \bigl\| \mathbf{V}_{\text{succ}}^{(\ell)} - \mathbf{V}_{\text{fail}}^{(\ell)} \bigr\|_2 .
\end{equation}
The rightmost panels of Figure~\ref{fig:layerwise-visualization} plot $d^{(\ell)}$ across layers. We observe a consistent upward trend, with the distance typically peaking in the final layers, aligning with the PCA visualizations and indicating that deeper representations encode more explicit and separable correctness signals.

\section{Conclusion}
In this work, we have demonstrated that the internal reasoning process of an LLM is not an inscrutable black box but a geometrically structured space containing clear, accessible signals of correctness. Our non-parametric framework, \textsc{Clue}, validates this principle by achieving remarkable verification performance through simple geometric clustering of past experiences, proving more robust than both LLM-judges and confidence-based methods. Critically, we uncover a fundamental connection between a model's training paradigm and its internal geometry: models fine-tuned with Reinforcement Learning develop cleanly separable representations for correct and incorrect reasoning, a property largely absent in their SFT- counterparts. This insight suggests a paradigm shift for the field, moving beyond the evaluation of final outputs towards the direct analysis and shaping of the reasoning process itself. We believe that the success of our minimalist approach opens the door to developing a new class of lightweight, generalizable verifiers and inspires novel training objectives that explicitly optimize for internal representational clarity.

\bibliography{colm2024_conference}

\begin{thebibliography}{40}
\providecommand{\natexlab}[1]{#1}
\providecommand{\url}[1]{\texttt{#1}}
\expandafter\ifx\csname urlstyle\endcsname\relax
  \providecommand{\doi}[1]{doi: #1}\else
  \providecommand{\doi}{doi: \begingroup \urlstyle{rm}\Url}\fi

\bibitem[An et~al.(2025)An, Xie, Li, Li, Zhang, Gong, Zhong, Xu, Qiu, Wang, and Kong]{Polaris2025}
Chenxin An, Zhihui Xie, Xiaonan Li, Lei Li, Jun Zhang, Shansan Gong, Ming Zhong, Jingjing Xu, Xipeng Qiu, Mingxuan Wang, and Lingpeng Kong.
\newblock Polaris: A post-training recipe for scaling reinforcement learning on advanced reasoning models, 2025.
\newblock URL \url{https://hkunlp.github.io/blog/2025/Polaris}.

\bibitem[Bai et~al.(2022)Bai, Kadavath, Kundu, Askell, Kernion, Jones, Chen, Goldie, Mirhoseini, McKinnon, et~al.]{bai2022constitutional}
Yuntao Bai, Saurav Kadavath, Sandipan Kundu, Amanda Askell, Jackson Kernion, Andy Jones, Anna Chen, Anna Goldie, Azalia Mirhoseini, Cameron McKinnon, et~al.
\newblock Constitutional ai: Harmlessness from ai feedback.
\newblock \emph{arXiv preprint arXiv:2212.08073}, 2022.

\bibitem[Bartoszcze et~al.(2025)Bartoszcze, Munshi, Sukidi, Yen, Yang, Williams-King, Le, Asuzu, and Maple]{bartoszcze2025representationengineeringlargelanguagemodels}
Lukasz Bartoszcze, Sarthak Munshi, Bryan Sukidi, Jennifer Yen, Zejia Yang, David Williams-King, Linh Le, Kosi Asuzu, and Carsten Maple.
\newblock Representation engineering for large-language models: Survey and research challenges, 2025.

\bibitem[Belrose et~al.(2023)Belrose, Furman, Smith, Halawi, Ostrovsky, McKinney, Biderman, and Steinhardt]{belrose2023eliciting}
Nora Belrose, Zach Furman, Logan Smith, Danny Halawi, Igor Ostrovsky, Lev McKinney, Stella Biderman, and Jacob Steinhardt.
\newblock Eliciting latent predictions from transformers with the tuned lens.
\newblock \emph{arXiv preprint arXiv:2303.08112}, 2023.

\bibitem[Chen et~al.(2025)Chen, Arditi, Sleight, Evans, and Lindsey]{chen2025personavectorsmonitoringcontrolling}
Runjin Chen, Andy Arditi, Henry Sleight, Owain Evans, and Jack Lindsey.
\newblock Persona vectors: Monitoring and controlling character traits in language models, 2025.

\bibitem[Cobbe et~al.(2021)Cobbe, Kosaraju, Bavarian, Chen, Jun, Kaiser, Plappert, Tworek, Hilton, Nakano, et~al.]{cobbe2021training}
Karl Cobbe, Vineet Kosaraju, Mohammad Bavarian, Mark Chen, Heewoo Jun, Lukasz Kaiser, Matthias Plappert, Jerry Tworek, Jacob Hilton, Reiichiro Nakano, et~al.
\newblock Training verifiers to solve math word problems.
\newblock \emph{arXiv preprint arXiv:2110.14168}, 2021.

\bibitem[Fu et~al.(2025{\natexlab{a}})Fu, Conde, Mart{\'\i}nez, Grandury, and Reviriego]{fu2025multiple}
Tairan Fu, Javier Conde, Gonzalo Mart{\'\i}nez, Mar{\'\i}a Grandury, and Pedro Reviriego.
\newblock Multiple choice questions: Reasoning makes large language models (llms) more self-confident even when they are wrong.
\newblock \emph{arXiv preprint arXiv:2501.09775}, 2025{\natexlab{a}}.

\bibitem[Fu et~al.(2025{\natexlab{b}})Fu, Wang, Tian, and Zhao]{fu2025deep}
Yichao Fu, Xuewei Wang, Yuandong Tian, and Jiawei Zhao.
\newblock Deep think with confidence.
\newblock \emph{arXiv preprint arXiv:2508.15260}, 2025{\natexlab{b}}.

\bibitem[Geng et~al.(2023)Geng, Cai, Wang, Koeppl, Nakov, and Gurevych]{geng2023survey}
Jiahui Geng, Fengyu Cai, Yuxia Wang, Heinz Koeppl, Preslav Nakov, and Iryna Gurevych.
\newblock A survey of confidence estimation and calibration in large language models.
\newblock \emph{arXiv preprint arXiv:2311.08298}, 2023.

\bibitem[Glaese et~al.(2022)Glaese, McAleese, Tr{\k{e}}bacz, Aslanides, Firoiu, Ewalds, Rauh, Weidinger, Chadwick, Thacker, et~al.]{glaese2022improving}
Amelia Glaese, Nat McAleese, Maja Tr{\k{e}}bacz, John Aslanides, Vlad Firoiu, Timo Ewalds, Maribeth Rauh, Laura Weidinger, Martin Chadwick, Phoebe Thacker, et~al.
\newblock Improving alignment of dialogue agents via targeted human judgements.
\newblock \emph{arXiv preprint arXiv:2209.14375}, 2022.

\bibitem[Guo et~al.(2025{\natexlab{a}})Guo, Yang, Zhang, Song, Zhang, Xu, Zhu, Ma, Wang, Bi, et~al.]{guo2025deepseek}
Daya Guo, Dejian Yang, Haowei Zhang, Junxiao Song, Ruoyu Zhang, Runxin Xu, Qihao Zhu, Shirong Ma, Peiyi Wang, Xiao Bi, et~al.
\newblock Deepseek-r1: Incentivizing reasoning capability in llms via reinforcement learning.
\newblock \emph{arXiv preprint arXiv:2501.12948}, 2025{\natexlab{a}}.

\bibitem[Guo et~al.(2025{\natexlab{b}})Guo, Wu, Yang, and Yu]{guo2025miningintrinsicrewardsllm}
Jizhou Guo, Zhaomin Wu, Hanchen Yang, and Philip~S. Yu.
\newblock Mining intrinsic rewards from llm hidden states for efficient best-of-n sampling, 2025{\natexlab{b}}.

\bibitem[Hao et~al.(2024)Hao, Sukhbaatar, Su, Li, Hu, Weston, and Tian]{hao2024training}
Shibo Hao, Sainbayar Sukhbaatar, DiJia Su, Xian Li, Zhiting Hu, Jason Weston, and Yuandong Tian.
\newblock Training large language models to reason in a continuous latent space.
\newblock \emph{arXiv preprint arXiv:2412.06769}, 2024.

\bibitem[Hendrycks et~al.(2021)Hendrycks, Burns, Kadavath, Arora, Basart, Tang, Song, and Steinhardt]{hendrycks2021measuring}
Dan Hendrycks, Collin Burns, Saurav Kadavath, Akul Arora, Steven Basart, Eric Tang, Dawn Song, and Jacob Steinhardt.
\newblock Measuring mathematical problem solving with the math dataset.
\newblock \emph{arXiv preprint arXiv:2103.03874}, 2021.

\bibitem[Hosseini et~al.(2024)Hosseini, Yuan, Malkin, Courville, Sordoni, and Agarwal]{hosseini2024v}
Arian Hosseini, Xingdi Yuan, Nikolay Malkin, Aaron Courville, Alessandro Sordoni, and Rishabh Agarwal.
\newblock V-star: Training verifiers for self-taught reasoners.
\newblock \emph{arXiv preprint arXiv:2402.06457}, 2024.

\bibitem[Hurst et~al.(2024)Hurst, Lerer, Goucher, Perelman, Ramesh, Clark, Ostrow, Welihinda, Hayes, Radford, et~al.]{hurst2024gpt}
Aaron Hurst, Adam Lerer, Adam~P Goucher, Adam Perelman, Aditya Ramesh, Aidan Clark, AJ~Ostrow, Akila Welihinda, Alan Hayes, Alec Radford, et~al.
\newblock Gpt-4o system card.
\newblock \emph{arXiv preprint arXiv:2410.21276}, 2024.

\bibitem[Kadavath et~al.(2022{\natexlab{a}})Kadavath, Conerly, Askell, Henighan, Drain, Perez, Schiefer, Hatfield-Dodds, DasSarma, Tran-Johnson, Johnston, El-Showk, Jones, Elhage, Hume, Chen, Bai, Bowman, Fort, Ganguli, Hernandez, Jacobson, Kernion, Kravec, Lovitt, Ndousse, Olsson, Ringer, Amodei, Brown, Clark, Joseph, Mann, McCandlish, Olah, and Kaplan]{kadavath2022languagemodelsmostlyknow}
Saurav Kadavath, Tom Conerly, Amanda Askell, Tom Henighan, Dawn Drain, Ethan Perez, Nicholas Schiefer, Zac Hatfield-Dodds, Nova DasSarma, Eli Tran-Johnson, Scott Johnston, Sheer El-Showk, Andy Jones, Nelson Elhage, Tristan Hume, Anna Chen, Yuntao Bai, Sam Bowman, Stanislav Fort, Deep Ganguli, Danny Hernandez, Josh Jacobson, Jackson Kernion, Shauna Kravec, Liane Lovitt, Kamal Ndousse, Catherine Olsson, Sam Ringer, Dario Amodei, Tom Brown, Jack Clark, Nicholas Joseph, Ben Mann, Sam McCandlish, Chris Olah, and Jared Kaplan.
\newblock Language models (mostly) know what they know, 2022{\natexlab{a}}.

\bibitem[Kadavath et~al.(2022{\natexlab{b}})Kadavath, Conerly, Askell, Henighan, Drain, Perez, Schiefer, Hatfield-Dodds, DasSarma, Tran-Johnson, et~al.]{kadavath2022language}
Saurav Kadavath, Tom Conerly, Amanda Askell, Tom Henighan, Dawn Drain, Ethan Perez, Nicholas Schiefer, Zac Hatfield-Dodds, Nova DasSarma, Eli Tran-Johnson, et~al.
\newblock Language models (mostly) know what they know.
\newblock \emph{arXiv preprint arXiv:2207.05221}, 2022{\natexlab{b}}.

\bibitem[Knappe et~al.(2024)Knappe, Li, Chauhan, Chhua, Zhu, and O'Brien]{knappe2024semantic}
Tim Knappe, Ryan Li, Ayush Chauhan, Kaylee Chhua, Kevin Zhu, and Sean O'Brien.
\newblock Semantic self-consistency: Enhancing language model reasoning via semantic weighting.
\newblock \emph{arXiv preprint arXiv:2410.07839}, 2024.

\bibitem[Lightman et~al.(2023)Lightman, Kosaraju, Burda, Edwards, Baker, Lee, Leike, Schulman, Sutskever, and Cobbe]{lightman2023let}
Hunter Lightman, Vineet Kosaraju, Yuri Burda, Harrison Edwards, Bowen Baker, Teddy Lee, Jan Leike, John Schulman, Ilya Sutskever, and Karl Cobbe.
\newblock Let's verify step by step.
\newblock In \emph{The Twelfth International Conference on Learning Representations}, 2023.

\bibitem[Lin et~al.(2023)Lin, Trivedi, and Sun]{lin2023generating}
Zhen Lin, Shubhendu Trivedi, and Jimeng Sun.
\newblock Generating with confidence: Uncertainty quantification for black-box large language models.
\newblock \emph{arXiv preprint arXiv:2305.19187}, 2023.

\bibitem[Liu et~al.(2025)Liu, Diao, Lu, Hu, Dong, Choi, Kautz, and Dong]{liu2025prorl}
Mingjie Liu, Shizhe Diao, Ximing Lu, Jian Hu, Xin Dong, Yejin Choi, Jan Kautz, and Yi~Dong.
\newblock Prorl: Prolonged reinforcement learning expands reasoning boundaries in large language models.
\newblock \emph{arXiv preprint arXiv:2505.24864}, 2025.

\bibitem[Liu et~al.(2024)Liu, Ye, Xing, and Zou]{liu2024incontextvectorsmakingcontext}
Sheng Liu, Haotian Ye, Lei Xing, and James Zou.
\newblock In-context vectors: Making in context learning more effective and controllable through latent space steering, 2024.

\bibitem[Luo et~al.(2024)Luo, Liu, Liu, Phatale, Guo, Lara, Li, Shu, Zhu, Meng, Sun, and Rastogi]{luo2024improvemathematicalreasoninglanguage}
Liangchen Luo, Yinxiao Liu, Rosanne Liu, Samrat Phatale, Meiqi Guo, Harsh Lara, Yunxuan Li, Lei Shu, Yun Zhu, Lei Meng, Jiao Sun, and Abhinav Rastogi.
\newblock Improve mathematical reasoning in language models by automated process supervision, 2024.

\bibitem[Ma et~al.(2025)Ma, Liu, Jiang, Zhang, Ma, and Chen]{ma2025general}
Xueguang Ma, Qian Liu, Dongfu Jiang, Ge~Zhang, Zejun Ma, and Wenhu Chen.
\newblock General-reasoner: Advancing llm reasoning across all domains.
\newblock \emph{arXiv preprint arXiv:2505.14652}, 2025.

\bibitem[Madaan et~al.(2023)Madaan, Tandon, Gupta, Hallinan, Gao, Wiegreffe, Alon, Dziri, Prabhumoye, Yang, et~al.]{madaan2023self}
Aman Madaan, Niket Tandon, Prakhar Gupta, Skyler Hallinan, Luyu Gao, Sarah Wiegreffe, Uri Alon, Nouha Dziri, Shrimai Prabhumoye, Yiming Yang, et~al.
\newblock Self-refine: Iterative refinement with self-feedback.
\newblock \emph{Advances in Neural Information Processing Systems}, 36:\penalty0 46534--46594, 2023.

\bibitem[nostalgebraist(2020)]{nostalgebraist2020interpreting}
nostalgebraist.
\newblock interpreting gpt: the logit lens, August31 2020.
\newblock URL \url{https://www.lesswrong.com/posts/AcKRB8wDpdaN6v6ru/interpreting-gpt-the-logit-lens}.
\newblock LessWrong post.

\bibitem[Ouyang et~al.(2022)Ouyang, Wu, Jiang, Almeida, Wainwright, Mishkin, Zhang, Agarwal, Slama, Ray, et~al.]{ouyang2022training}
Long Ouyang, Jeffrey Wu, Xu~Jiang, Diogo Almeida, Carroll Wainwright, Pamela Mishkin, Chong Zhang, Sandhini Agarwal, Katarina Slama, Alex Ray, et~al.
\newblock Training language models to follow instructions with human feedback.
\newblock \emph{Advances in neural information processing systems}, 35:\penalty0 27730--27744, 2022.

\bibitem[Rein et~al.(2024)Rein, Hou, Stickland, Petty, Pang, Dirani, Michael, and Bowman]{rein2024gpqa}
David Rein, Betty~Li Hou, Asa~Cooper Stickland, Jackson Petty, Richard~Yuanzhe Pang, Julien Dirani, Julian Michael, and Samuel~R Bowman.
\newblock Gpqa: A graduate-level google-proof q\&a benchmark.
\newblock In \emph{First Conference on Language Modeling}, 2024.

\bibitem[Shen et~al.(2025)Shen, Wang, Shi, Wang, Zhao, and Gu]{shen2025efficient}
Xuan Shen, Yizhou Wang, Xiangxi Shi, Yanzhi Wang, Pu~Zhao, and Jiuxiang Gu.
\newblock Efficient reasoning with hidden thinking.
\newblock \emph{arXiv preprint arXiv:2501.19201}, 2025.

\bibitem[Snell et~al.(2024)Snell, Lee, Xu, and Kumar]{snell2024scaling}
Charlie Snell, Jaehoon Lee, Kelvin Xu, and Aviral Kumar.
\newblock Scaling llm test-time compute optimally can be more effective than scaling model parameters.
\newblock \emph{arXiv preprint arXiv:2408.03314}, 2024.

\bibitem[Tomaz et~al.(2025)Tomaz, Rosenblatt, Jones, and de~Lucena]{tomaz2025momentum}
Lorenzo Tomaz, Judd Rosenblatt, Thomas~Berry Jones, and Diogo~Schwerz de~Lucena.
\newblock Momentum point-perplexity mechanics in large language models.
\newblock \emph{arXiv preprint arXiv:2508.08492}, 2025.

\bibitem[Veeraboina(2023)]{aime_1983_2024}
Hemish Veeraboina.
\newblock Aime problem set 1983-2024, 2023.
\newblock URL \url{https://www.kaggle.com/datasets/hemishveeraboina/aime-problem-set-1983-2024}.

\bibitem[Wang et~al.(2022)Wang, Wei, Schuurmans, Le, Chi, Narang, Chowdhery, and Zhou]{wang2022self}
Xuezhi Wang, Jason Wei, Dale Schuurmans, Quoc Le, Ed~Chi, Sharan Narang, Aakanksha Chowdhery, and Denny Zhou.
\newblock Self-consistency improves chain of thought reasoning in language models.
\newblock \emph{arXiv preprint arXiv:2203.11171}, 2022.

\bibitem[Xiong et~al.(2024)Xiong, Santilli, Kirchhof, Golinski, and Williamson]{xiong2024efficient}
Miao Xiong, Andrea Santilli, Michael Kirchhof, Adam Golinski, and Sinead Williamson.
\newblock Efficient and effective uncertainty quantification for llms.
\newblock In \emph{Neurips Safe Generative AI Workshop 2024}, 2024.

\bibitem[Yang et~al.(2025{\natexlab{a}})Yang, Li, Yang, Zhang, Hui, Zheng, Yu, Gao, Huang, Lv, et~al.]{yang2025qwen3}
An~Yang, Anfeng Li, Baosong Yang, Beichen Zhang, Binyuan Hui, Bo~Zheng, Bowen Yu, Chang Gao, Chengen Huang, Chenxu Lv, et~al.
\newblock Qwen3 technical report.
\newblock \emph{arXiv preprint arXiv:2505.09388}, 2025{\natexlab{a}}.

\bibitem[Yang et~al.(2025{\natexlab{b}})Yang, Si, Duan, Zhu, Zhu, Li, Lin, Cao, and Wang]{yang2025dynamicearlyexitreasoning}
Chenxu Yang, Qingyi Si, Yongjie Duan, Zheliang Zhu, Chenyu Zhu, Qiaowei Li, Zheng Lin, Li~Cao, and Weiping Wang.
\newblock Dynamic early exit in reasoning models, 2025{\natexlab{b}}.

\bibitem[Yao et~al.(2023)Yao, Yu, Zhao, Shafran, Griffiths, Cao, and Narasimhan]{yao2023tree}
Shunyu Yao, Dian Yu, Jeffrey Zhao, Izhak Shafran, Tom Griffiths, Yuan Cao, and Karthik Narasimhan.
\newblock Tree of thoughts: Deliberate problem solving with large language models.
\newblock \emph{Advances in neural information processing systems}, 36:\penalty0 11809--11822, 2023.

\bibitem[Zhang et~al.(2025)Zhang, Chen, Pan, Zhao, Panda, Li, and He]{zhang2025reasoningmodelsknowtheyre}
Anqi Zhang, Yulin Chen, Jane Pan, Chen Zhao, Aurojit Panda, Jinyang Li, and He~He.
\newblock Reasoning models know when they're right: Probing hidden states for self-verification, 2025.

\bibitem[Zheng et~al.(2023)Zheng, Chiang, Sheng, Zhuang, Wu, Zhuang, Lin, Li, Li, Xing, et~al.]{zheng2023judging}
Lianmin Zheng, Wei-Lin Chiang, Ying Sheng, Siyuan Zhuang, Zhanghao Wu, Yonghao Zhuang, Zi~Lin, Zhuohan Li, Dacheng Li, Eric Xing, et~al.
\newblock Judging llm-as-a-judge with mt-bench and chatbot arena.
\newblock \emph{Advances in neural information processing systems}, 36:\penalty0 46595--46623, 2023.

\end{thebibliography}
\bibliographystyle{colm2024_conference}

\newpage
\appendix

\section{Appendix}

\subsection{Algorithmic Details}
For completeness, we provide pseudocode for the two phases of \textsc{Clue}: the one-time centroid aggregation (Algorithm~\ref{alg:learning}) and the inference-time verification (Algorithm~\ref{alg:inference}). Both phases operate on \emph{activation-delta matrices} (\S\ref{sec:delta}) and use the \emph{layer-averaged Euclidean distance} in Eq.~\eqref{eq:layeravg}. The underlying LLM remains frozen throughout.

Algorithm~\ref{alg:learning} constructs the reference centroids from a labeled set of trajectories. We first partition the dataset by ground-truth labels. For each trajectory, we extract the hidden-state matrices at the boundaries of the explicit reasoning block (\texttt{<think>} and \texttt{</think>}) and compute the activation delta $\Delta \mathbf{h}_i\in\mathbb{R}^{L\times D}$ via Eq.~\eqref{eq:delta}. We then compute the element-wise mean within each class to obtain the success and failure \emph{centroid matrices} $\mathbf{V}_{\text{succ}}$ and $\mathbf{V}_{\text{fail}}$ (Eq.~\eqref{eq:centroids}), which serve as geometric references during inference.

\begin{algorithm}[H]
\caption{Constructing \textsc{Clue} Centroids (Learning Phase)}
\label{alg:learning}
\begin{algorithmic}[1]
\Require Labeled dataset $\mathcal{D}=\{(T_i,y_i)\}_{i=1}^N$
\State Initialize empty lists $\mathcal{H}_{\text{succ}},\,\mathcal{H}_{\text{fail}}$
\State Define index sets $\mathcal{I}_{\text{succ}}=\{\,i\mid y_i=1\,\}$, $\mathcal{I}_{\text{fail}}=\{\,i\mid y_i=0\,\}$
\For{each $i\in\mathcal{I}_{\text{succ}}$}
    \State Extract $\mathbf{h}_{\text{start},i}\in\mathbb{R}^{L\times D}$ and $\mathbf{h}_{\text{end},i}\in\mathbb{R}^{L\times D}$
    \State Compute $\Delta \mathbf{h}_i \gets \mathbf{h}_{\text{end},i}-\mathbf{h}_{\text{start},i}$ \hfill (Eq.~\ref{eq:delta})
    \State Append $\Delta \mathbf{h}_i$ to $\mathcal{H}_{\text{succ}}$
\EndFor
\For{each $i\in\mathcal{I}_{\text{fail}}$}
    \State Extract $\mathbf{h}_{\text{start},i}$ and $\mathbf{h}_{\text{end},i}$
    \State Compute $\Delta \mathbf{h}_i \gets \mathbf{h}_{\text{end},i}-\mathbf{h}_{\text{start},i}$
    \State Append $\Delta \mathbf{h}_i$ to $\mathcal{H}_{\text{fail}}$
\EndFor
\State $\mathbf{V}_{\text{succ}} \gets \text{mean}(\mathcal{H}_{\text{succ}})$ \hfill (Eq.~\ref{eq:centroids})
\State $\mathbf{V}_{\text{fail}} \gets \text{mean}(\mathcal{H}_{\text{fail}})$ \hfill (Eq.~\ref{eq:centroids})
\State \Return $\mathbf{V}_{\text{succ}},\,\mathbf{V}_{\text{fail}}$
\end{algorithmic}
\end{algorithm}

Algorithm~\ref{alg:inference} describes the inference procedure. Given a new trajectory $T_{\text{new}}$, we compute its activation delta $\Delta \mathbf{h}_{\text{new}}$ as in Eq.~\eqref{eq:delta}, measure its distances to the two centroid matrices using Eq.~\eqref{eq:layeravg}, and decide by nearest centroid.

\begin{algorithm}[H]
\caption{Verification with \textsc{Clue} (Inference Phase)}
\label{alg:inference}
\begin{algorithmic}[1]
\Require New trajectory $T_{\text{new}}$; centroids $\mathbf{V}_{\text{succ}},\,\mathbf{V}_{\text{fail}}$
\State Extract $\mathbf{h}_{\text{start,new}}$ and $\mathbf{h}_{\text{end,new}}$ from $T_{\text{new}}$
\State Compute $\Delta \mathbf{h}_{\text{new}} \gets \mathbf{h}_{\text{end,new}}-\mathbf{h}_{\text{start,new}}$ \hfill (Eq.~\ref{eq:delta})
\State $d_{\text{succ}} \gets d(\Delta \mathbf{h}_{\text{new}},\,\mathbf{V}_{\text{succ}})$ \hfill (Eq.~\ref{eq:layeravg})
\State $d_{\text{fail}} \gets d(\Delta \mathbf{h}_{\text{new}},\,\mathbf{V}_{\text{fail}})$ \hfill (Eq.~\ref{eq:layeravg})
\If{$d_{\text{succ}} < d_{\text{fail}}$}
    \State \Return $1$ \Comment{classified as correct}
\Else
    \State \Return $0$ \Comment{classified as incorrect}
\EndIf
\end{algorithmic}
\end{algorithm}

\end{document}